\documentclass[10pt,twocolumn,letterpaper]{article}

\usepackage[pagenumbers]{cvpr} 

\usepackage{xcolor}         
\definecolor{cvprblue}{rgb}{0.21,0.49,0.74}
\usepackage[pagebackref,breaklinks,colorlinks,allcolors=cvprblue]{hyperref}

\usepackage[utf8]{inputenc} 
\usepackage[T1]{fontenc}    
\usepackage{url}            
\usepackage{booktabs}       
\usepackage{amsfonts}       
\usepackage{nicefrac}       
\usepackage{microtype}      
\usepackage{xspace}
\usepackage{makecell}
\usepackage[acronym]{glossaries}
\usepackage{multirow}
\usepackage{graphicx,wrapfig,lipsum}
\usepackage{algorithm}      
\usepackage{algpseudocode}
\usepackage{mathtools}
\usepackage{nicematrix}
\usepackage{booktabs}
\usepackage{pifont}

\robustify{\gls}

\newacronym{DNN}{DNN}{Deep Neural Networks}
\newacronym{DL}{DL}{Deep Learning}
\newacronym{CNN}{CNN}{Convolutional Neural Network}
\newacronym{ML}{ML}{Machine Learning}
\newacronym{NN}{NN}{Neural Network}
\newacronym{ViT}{ViT}{Vision Transformer}
\newacronym{MHA}{MHA}{Multi-head Attention}
\newacronym{MLP}{MLP}{Multilayer Perceptron}
\newacronym{NLP}{NLP}{Natural Language Processing}
\newacronym{NORM}{NORM}{Normalization Layer}
\newacronym{EMA}{EMA}{Exponential Moving Average}
\newacronym{MoE}{MoE}{Mixture of Experts}
\newacronym{LLM}{LLM}{Large Language Model}

\def\paperID{33591} 
\def\confName{CVPR}
\def\confYear{2026}
\newcommand{\nameNoSpace}{\textsl{ThinkingViT}} 
\newcommand{\name}{\nameNoSpace\xspace}

\title{\name: Matryoshka Thinking Vision Transformer for Elastic Inference}

{\small
\author{
Ali Hojjat\(^{1,2}\),
Janek Haberer\(^{1}\),
Sören Pirk\(^{1}\),
Olaf Landsiedel\(^{2,1,3}\)\\[-0.2em]
{\footnotesize
\(^{1}\)~Kiel University, Germany \quad \(^{2}\)~Hamburg University of Technology (TUHH), Germany \quad \(^{3}\)~UNU-INWEH, Germany
}\\[-0.25em]
{\footnotesize
\texttt{\{ali.hojjat,\, olaf.landsiedel\}@tuhh.de} \quad
\texttt{\{janek.haberer,\, soeren.pirk\}@cs.uni-kiel.de}
}\\[-0.35em]
{\scriptsize
\href{https://github.com/ds-kiel/ThinkingViT}{\texttt{github.com/ds-kiel/ThinkingViT}}
}
}
}
\begin{document}
\maketitle

\begin{abstract}
\glspl{ViT} deliver SOTA performance, yet their fixed computational budget prevents scalable deployment across heterogeneous hardware.
Recent Matryoshka-style Transformer architectures mitigate this by embedding nested subnetworks within a single model to enable scalable inference. However, these models allocate the same amount of compute to all inputs, regardless of their complexity, which leads to inefficiencies.
To address this, we introduce \nameNoSpace, a nested \gls{ViT} architecture that employs progressive thinking stages to dynamically adjust inference computation based on input difficulty.
\name first activates a small subset of the most important attention heads to produce an initial prediction. If the prediction confidence exceeds a predefined threshold, inference terminates early.
Otherwise, within the same backbone, it activates a larger subset of attention heads and conducts a new forward pass. This process continues iteratively until the model reaches the predefined confidence level or exhausts its maximum capacity.
To boost the performance of subsequent rounds, we introduce a Token Recycling approach that fuses the input embeddings with the embeddings from the previous stage.
Experiments show that \name surpasses nested baselines by up to 2.0 percentage points (p.p.) in accuracy at the same throughput and by up to 2.9 p.p. at equal GMACs on ImageNet-1K.
We show that the backbone-preserving design of \name allows it to serve as a plug-in upgrade for \glspl{ViT} in downstream tasks such as semantic segmentation. We also demonstrate that \name transfers effectively to other architectures such as Swin Transformers.
\end{abstract}

\begin{figure}[!b]
    \centering
    \begin{subfigure}{0.49\textwidth}
        \centering
        \includegraphics[width=0.82\textwidth]{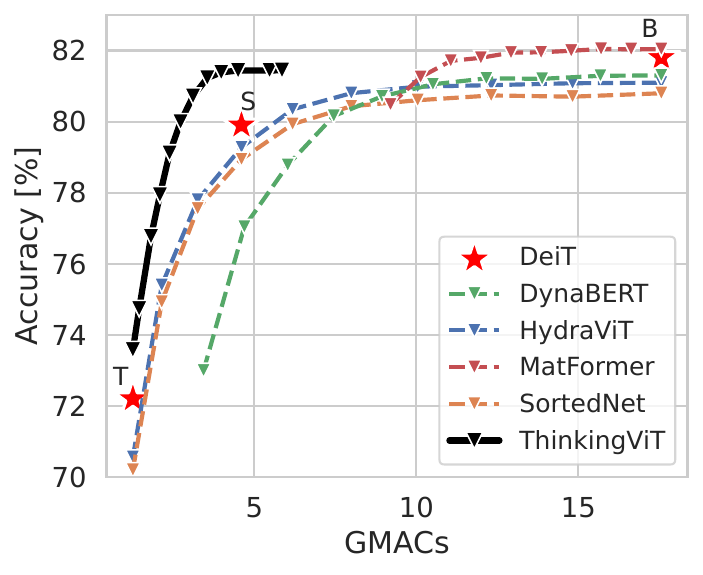}
        \label{fig:gmacs_imagenet_val}
        \caption{GMACs vs. Accuracy on ImageNet-1K}
    \end{subfigure}
    \hfill
    \begin{subfigure}{0.49\textwidth}
        \centering
        \includegraphics[width=0.82\textwidth]{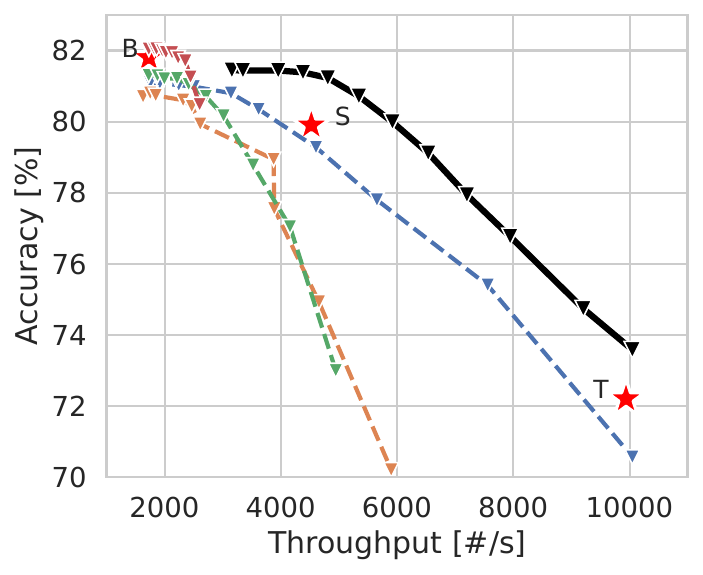}
        \label{fig:throughput_imagenet_val}
        \caption{Throughput vs. Accuracy on ImageNet-1K}
    \end{subfigure}
    \caption{Comparison of \name with MatFormer{\tiny (NeurIPS'24)}~\citep{devvrit2024matformer}, HydraViT{\tiny (NeurIPS'24)}~\citep{haberer2024hydravit}, SortedNet{\tiny (NeurIPS'23-W)}~\citep{valipour2023sortednet}, and DynaBERT{\tiny (NeurIPS'20)}~\citep{hou2020dynabert}, evaluated in terms of GMACs and throughput on an A100.
    All baselines have a dynamic width within a standard \gls{ViT} backbone and are trained following the training recipes in~\cite{touvron2021training}. \name is trained with two progressive thinking stages using 3 and 6 heads, and consistently surpasses baselines by up to 2.0 p.p. at the same throughput and by up to 2.9 p.p. at the same GMACs. See Appendix~\ref{appendix_gmacs_imagenet_val_for_small} for a comparison with smaller baseline models.}
    \label{fig:gmacs_throughput_val}
\label{fig:imagenet_val}
\end{figure}

\begin{figure*}
  \centering
   \includegraphics[trim=2 0 0 0, clip, width=1.\textwidth]{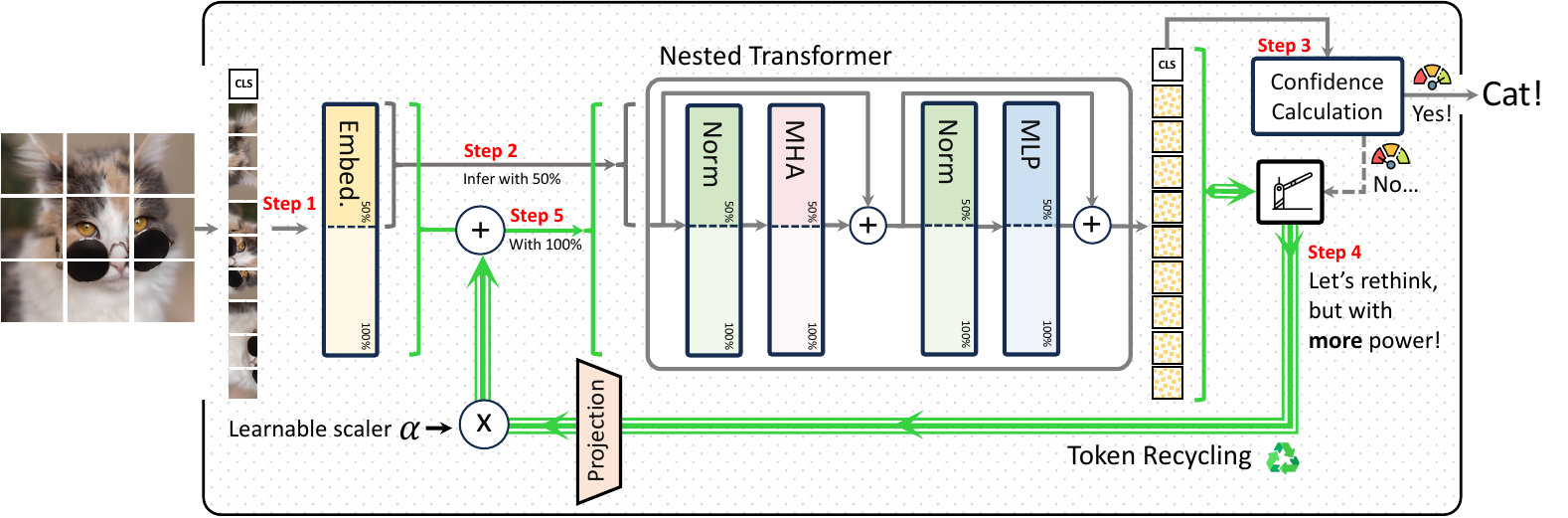}
\caption{\textbf{Nested progressive inference with Token Recycling in \name.}
After embedding the input, \name first activates a subset of the model, including the first attention heads (e.g., 50\%), to produce an initial prediction. Due to the training procedure, these heads capture the most important features.
If the certainty exceeds a threshold (easy inputs), inference terminates early to save computation. Otherwise, the resulting tokens are fused back into the input via a projection and a learnable scaling factor \(\alpha\), which controls how much prior knowledge is recycled. The model then \emph{thinks more} by reprocessing the fused tokens using a larger subset of the attention heads (e.g., 100\%) for a refined prediction. \name enables elastic inference across different hardware budgets by adjusting the confidence threshold. The number of thinking stages and the proportion of attention heads activated at each stage can be flexibly configured based on efficiency and accuracy trade-offs, see Section~\ref{sec:trade_offs}.}
   \label{fig:demo}
\end{figure*}

\section{Introduction}

\textbf{Motivation:}
\acrfullpl{ViT} \citep{dosovitskiy2021an} have achieved state-of-the-art results across numerous image recognition tasks \citep{liu2022swin, hatamizadeh2024mambavision}, yet their non-elastic inference pipelines impose a uniform and often excessive computational cost. This lack of flexibility poses a significant limitation in real-world deployments, where devices vary widely in computational power and latency constraints \citep{xu2025edgevit}. From high-throughput servers to mobile and embedded platforms, modern applications require elastic inference \citep{cai2024flextron}, where a model can adjust its computational footprint to match the capabilities of the target hardware. Without such adaptability, \glspl{ViT} struggle to meet the performance and efficiency tradeoffs necessary for scalable and widespread deployment.

\textbf{Limitations of Prior Work:}
Recent progress in \emph{nested} Transformer architectures offers a compelling pathway toward elastic inference \citep{haberer2024hydravit, devvrit2024matformer, zhang2024slicing}. These methods embed multiple nested subnetworks within a single Transformer backbone, enabling dynamic subnetwork selection at inference time. This strategy allows models to operate under varying latency and hardware constraints without any extra tuning or post-training adjustments, all while maintaining a unified parameter set.
However, these approaches typically allocate a \emph{fixed} computational budget per input, missing the opportunity to distinguish between simple and complex samples, which leads to inefficiencies, especially under tight resource constraints.

\textbf{Image-Based Routing:}
To make nested models adaptable, we must route the input image to the appropriate subnetwork based on its difficulty. While the concept of routing appears in \gls{MoE} models \citep{zhou2022mixture}, we cannot use them out of the box since these routers operate at the token level by directing token embeddings via lightweight \glspl{MLP}, which struggle to capture the global complexity of the image. In practice, accurately estimating image difficulty demands representational power comparable to that of a full-scale classifier, a requirement that such compact routers cannot meet \citep{ong2025routellm, ding2025occam}.
Moreover, dedicating substantial compute to a separate router that merely forwards the input to another model introduces overhead.
This leads to a critical design question:
\textit{How can we enable input-aware compute allocation within a \textbf{nested} Vision Transformer, without relying on a separate costly routing mechanism?}

\textbf{\name:}
Inspired by recent advances in thinking-based \glspl{LLM} \citep{guo2025deepseek, shao2024deepseekmath}, we introduce \name, a \gls{ViT} built upon the vanilla \gls{ViT} architecture \citep{dosovitskiy2021an} that dynamically adjusts its inference effort based on input complexity through \emph{nested progressive thinking stages}. \name begins by activating a subset of the nested network, including the most important attention heads (e.g., the top 50\%) to generate an initial prediction along with a certainty score. If the certainty is sufficiently high (the "aha" moment \citep{guo2025deepseek}), inference terminates early.
Otherwise, the processed embeddings are fused with the original input embeddings and passed through a larger subset of the nested network, using an increased set of attention heads (e.g., 100\%) to perform a more thorough re-evaluation.
This token fusion mechanism allows the model to avoid thinking from scratch; instead, it \textbf{recycles} the knowledge gained in the first pass, refining its prediction with improved accuracy; see Figure~\ref{fig:demo}.
Our proposed progressive thinking mechanism enables the model not only to \emph{think more}, but also to \emph{think more powerfully} when processing ambiguous inputs, ensuring that more challenging examples receive greater representational capacity, i.e., more attention heads.
In this pipeline, the prediction certainty serves as the central mechanism that governs elastic inference: the model halts early for cases with high certainty and progressively deepens computation for more uncertain inputs. 
By adjusting the certainty threshold, the model naturally balances performance against efficiency, supporting elastic inference without requiring any separate routing mechanism.
Additionally, nesting all stages within a single unified model avoids parameter duplication and enables more efficient training and improved accuracy.

\begin{table}[t]
\centering
\small
\caption{Looping over the same model offers limited accuracy gains despite increased computation.}
\setlength{\tabcolsep}{4pt}
\begin{tabular}{lccccc}

    \toprule
    \textbf{Model} & \textbf{Depth} & \textbf{Size [M]} & \textbf{GMACs} & \textbf{Acc.} \\
    \toprule
    DeiT‑Tiny & 12 & 5.70  & 1.25 & 72.20 \\
    \midrule
    \multicolumn{5}{l}{\textbf{Naïve Iteration $\circlearrowright$}} \\
    + 1× iteration & 24 & 5.70  & 2.5 & 74.00 \\
    + 2× iteration & 36 & 5.70  & 3.75 & 74.10 \\
    + 3× iteration & 48 & 5.70  & 5.00 & 73.60 \\
    \bottomrule

\end{tabular}
\label{tab:naive_recursion}
\end{table}

Furthermore, unlike iterative refinement in \glspl{LLM} that repeatedly use the same model \citep{guo2025deepseek}, naïve iterative chains that simply refeed a \gls{ViT}’s own outputs into the same network fail to deliver consistent improvements and quickly saturate in vision models; see Table~\ref{tab:naive_recursion}.
These findings underscore the need for the progressive expansion strategy adopted in \name.

\noindent\textbf{Contributions:}
 

\begin{enumerate}

  \item We introduce \name, a thinking‑based \acrfull{ViT} that brings input adaptivity to nested Transformers by dynamically adapting computation based on image difficulty. 


   \item \name executes multiple rounds of inference by progressively activating larger subsets of attention heads, allowing the model to halt early for easy inputs based on the certainty of the predictions, while allocating greater capacity to harder examples that require richer representations.
  
  \item \name introduces Token Recycling to condition each subsequent round of inference on the features produced in the previous round, improving overall accuracy. 
  \item \name achieves up to 2.0 percentage points higher accuracy at equal throughput, and up to 2.9 points higher at the same GMACs compared to baseline models (see Figure~\ref{fig:gmacs_throughput_val}), and extends to hierarchical architectures and downstream tasks, confirming broad adaptability.
\end{enumerate}

\section{Related Work}



\textbf{Nested Models:} Slimmable networks first demonstrated that a single model can operate at multiple widths using shared weights and width-specific normalization~\citep{yu2018slimmable, yu2019universally}. Many subsequent works build on this idea.
For instance, MatFormer~\citep{devvrit2024matformer}, based on Matryoshka Representation Learning~\citep{kusupati2022matryoshka}, introduces multiple nested subsets within the hidden layer of \gls{MLP}.
DynaBERT~\citep{hou2020dynabert} slices the \gls{MHA} and \gls{MLP} layers, but does not slice embeddings across layers since it relies on knowledge distillation.
SortedNet \citep{valipour2023sortednet} generalizes nesting across \gls{MLP}, \gls{NORM}, \gls{MHA}, and embeddings, although it retains a fixed number of attention heads. 
HydraViT~\citep{haberer2024hydravit} and Slicing ViT~\citep{zhang2024slicing} enable slicing across embeddings, \gls{NORM}, \gls{MLP}, and \gls{MHA}, and support a dynamic number of heads.
However, like the other mentioned methods, they apply a fixed compute budget per input, limiting their ability to adapt based on input complexity and leading to inefficiencies under constrained resources.

\textbf{Routing:}
Routing mechanisms got popular in the \gls{MoE} framework~\citep{zhou2022mixture} and have since been extended to nested designs such as MoNE~\citep{jain2024mixture}, MoD~\citep{raposo2024mixture}, and AMoD~\citep{gadhikar2025attention}. Flextron introduces a surrogate loss predictor to guide token routing~\citep{cai2024flextron}. These methods typically rely on lightweight \glspl{MLP}, which lack the representational capacity to route at the image level, where decisions often demand the reasoning power of a full classifier. Ensemble-based strategies like Selective Query~\citep{kag2023efficient}, OCCAM~\citep{ding2025occam}, and RouteLLM~\citep{ong2025routellm} perform input-aware routing using full models as gates.
However, these approaches operate over a fixed set of pretrained models (e.g., CNNs or Transformers) and do not enable any knowledge transfer between the router and the routed models. In contrast, \name recycles tokens across stages, allowing each round to build on previous inferences for improved prediction.

\textbf{Thinking:}
\glspl{LLM} have introduced mechanisms for adaptive reasoning depth \citep{el2025competitive, jaech2024openai}. These methods often use reinforcement learning to dynamically adjust the number of reasoning steps based on input complexity \citep{guo2025deepseek, shao2024deepseekmath}. However, excessive reasoning length can inflate inference cost without proportional gains in accuracy \citep{kumar2025overthink}. Motivated by this, recent efforts seek to preserve the benefits of deep reasoning while avoiding redundant computation by introducing test-time adaptability or confidence-based early stopping \citep{muennighoff2025s1, zeng2025revisiting}.
These models, however, reuse the same network in each round, which, as reported in Table~\ref{tab:naive_recursion}, results in limited gains. In contrast, \name scales up its capacity during rethinking to overcome this challenge.

\textbf{Early-exit:} Early-exit methods reduce inference cost by attaching intermediate classifiers that let models stop once predictions are confident. Classic approaches such as BranchyNet~\cite{teerapittayanon2016branchynet}, MSDNet~\cite{huang2017multi}, and SDN~\cite{kaya2019shallow} introduced multi-exit architectures, later extended to Transformers with PABEE~\cite{zhou2020bert} and BERxiT~\cite{xin2021berxit}, and to ViTs with ViT-EE~\cite{bakhtiarnia2021multi}, PCEE~\cite{zhang2022pcee}, and LGViT~\cite{xu2023lgvit}. Jointly learned strategies such as JEI-DNN~\cite{regol2023jointly} further improve reliability by co-training gating and exit classifiers. While these methods offer dynamic depth, they rely on fixed-width models; in contrast, \name performs multiple loops over the model with progressively increased width.

\section{\name}
\label{sec:design}



\textbf{How \name works:}
After embedding the input, \name runs inference with a small subset of attention heads that, due to the training procedure, are the most important heads. The model stops if the classification confidence is high. Otherwise, within the same backbone, it activates a larger subset of attention heads and performs another inference step. This iterative expansion continues until the model reaches the predefined confidence or exhausts its maximum capacity. Token Recycling further improves later stages by fusing input embeddings with features from previous stages. By adjusting the confidence threshold, \name enables elastic inference across different target latencies.

\subsection{Nested Vision Transformer - Sorting Heads}  
\name is built on top of the standard \gls{ViT} architecture. Let $V_{D,H}$ be a \gls{ViT} model with $H$ attention heads and the embedding dimension of $D$. ViT begins by tokenizing the input image $x$ into $P$ non-overlapping patches, each of which is projected into a $D$-dimensional embedding vector $\mathcal{E}^{D}$ using a convolution. After adding positional encodings, the resulting sequence is processed by $L$ Transformer blocks, where $z_0 = \mathcal{E}^{D}$:
\begin{equation}
\quad z_l = \text{Block}_l(z_{l-1}) \quad \text{for } l = 1,\dots,L
\label{eq:vit_blocks}
\end{equation}

\noindent
\textbf{Inducing sorted subnetworks:}  
Inspired by~\cite{haberer2024hydravit}, \name induces $n$ sorted subnetworks within this architecture. Each subnetwork is denoted as $V_{d_i,h_i}$ and is constructed using the first $d_i$ embedding values of each token and the first $h_i$ attention heads from the full model. To build such a nested structure, \name slices all components of the \gls{ViT}, including the embedding layer, attention modules, \gls{MLP} blocks, and normalization layers, according to these indices. This yields contained subnetworks with progressively increasing embedding dimensionality and attention capacity:
\begin{align}
&V_{d_1,h_1}(x) \subset V_{d_2,h_2}(x) \subset \cdots \subset V_{d_n,h_n}(x), \\
&d_1 < d_2 < \cdots < d_n,\quad h_1 < h_2 < \cdots < h_n
\end{align}

\subsection{Looping nested Vision Transformers through Token Recycling}
\label{Token_Recycling}
Inspired by recent advances in reasoning models~\citep{guo2025deepseek, shao2024deepseekmath}, \name\ operates through multiple rounds of progressive refinement using our proposed "Token Recycling" mechanism. Although in practice, we find that two subnetworks are sufficient to achieve strong performance (as demonstrated in Figure~\ref{fig:variants}), we present the general multi-round formulation here for clarity and broader applicability.

After constructing the nested architecture, \name\ begins by using a small subnetwork $V_{d_i,h_i}$, with the embedding dimension of $d_i$ and $h_i$ attention heads. This subnetwork processes the input image to predict the class and generate token embeddings $z_{L}$. Subsequently, to refine its prediction, the model expands computational capacity by activating a larger subnetwork $V_{d_j, h_j}(x)$, featuring $h_j$ attention heads, where $h_i < h_j$ and $d_i < d_j$. Afterwards, it fuses the produced token embeddings $z_{L}$ with the new input embeddings $\mathcal{E}^{d_j}(x) \in \mathbb{R}^{P \times d_j}$, through a projection layer and scaled by a learnable parameter $\alpha$, determining the weight of embeddings "recalled" from the previous step:
\begin{equation}
\mathcal{E}^{d_{j}}_{\mathrm{fused}} = \alpha \cdot \mathrm{Proj}_{d_i \rightarrow d_j}(z_{L}) + \mathcal{E}^{d_{j}}(x),
\end{equation}

This \emph{Token Recycling} mechanism allows the model to reuse prior representations instead of reprocessing from scratch. The process continues iteratively and, unlike language models that reuse the same network for reasoning~\citep{guo2025deepseek}, our approach increases model capacity by activating progressively larger subsets of attention heads, enhancing representational capacity at each step. Progressive expansion is essential in vision tasks, where naïve recursion quickly plateaus, as shown in Table~\ref{tab:naive_recursion}. Appendix~\ref{appendix_fusing_ablation} and Appendix~\ref{appendix_token_recycling_ablation} compare recycling and fusion strategies in detail.

\subsection{Training all subnetworks jointly}  
\label{training_jointly}
During training, \name executes all $n$ thinking rounds and minimizes a weighted classification loss $\mathcal{L}_\text{cls}$ across all stages:
\begin{equation}
\mathcal{L} = \sum_{i=1}^n \lambda_i \cdot \mathcal{L}_\text{cls}(V_{d_i,h_i}(x), y)
\end{equation}
where $y$ is the ground-truth label and $ \lambda_i $ controls the contribution of each subnetwork to the global objective. 
For a small number of subnetworks, it is computationally feasible to optimize all subnetworks simultaneously, as the gradient computation graph remains reasonably compact. However, as $n$ increases, training all subnetworks jointly becomes computationally intensive. In such cases, we adopt strategies such as the sandwich rule~\citep{yu2019universally} and stochastic subnetwork sampling~\citep{haberer2024hydravit} to reduce overhead.

\subsection{The ``Aha!'' moment}

By inducing nested iterative subnetworks inside the model, \name builds a hierarchy of $n$ nested thinking steps, where each step operates on top of the previous one to refine the prediction. However, different inputs have different complexities, and for some inputs, the model becomes confident after only $k$ iterations, where $k < n$ (e.g., after a single round of thinking). The "aha!" moment occurs when the model recognizes that it has reached sufficient certainty, allowing it to allocate an appropriate number of thinking rounds based on the input complexity~\citep{guo2025deepseek}. Let $f_k$ denote the softmax output produced at the $k$-th iteration. After the $k^{th}$ round, the model measures its certainty using Shannon entropy:
\begin{equation}
\mathcal{H}(f_k) = -\sum_{c=1}^C f_k^{(c)} \log f_k^{(c)},
\end{equation}
where $C$ is the number of classes. If the entropy $\mathcal{H}(f_k)$ is lower than a predefined threshold $\tau$, inference halts early and the model accepts the current prediction. Otherwise, \name activates a larger subnetwork to further refine the result.
This simple metric performs on par with more complex criteria \citep{jitkrittum2023when} on ImageNet-1K. Our design also allows alternative routing modules \citep{kag2023efficient, ding2025occam} to be integrated as drop-in replacements without changing the architecture or training.

\subsection{Elastic inference}

\name supports elastic inference by using its own certainty to determine when to stop. The entropy threshold $\tau$ governs the trade-off between efficiency and accuracy. A low threshold enforces stricter confidence requirements, causing most inputs to undergo more thinking steps, which increases computational cost but improves accuracy. In contrast, a high threshold allows more inputs to exit early, reducing latency and computation cost at the expense of potential accuracy loss.
This flexibility enables users to tune the model at runtime according to specific resource and performance needs.


\begin{figure}[t]
    \centering
    \includegraphics[width=.8\linewidth]{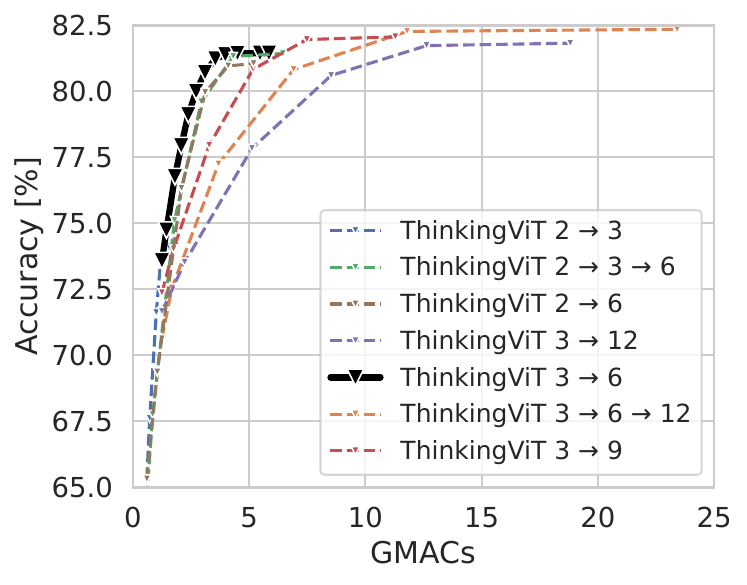}
    \caption{Accuracy vs. GMACs for \name variants on ImageNet-1K.
    3H~$\rightarrow$~6H achieves the best trade-off, while 2H~$\rightarrow$~3H~$\rightarrow$~6H covers the widest range with only a slight accuracy drop. See Appendix~\ref{sec:thinking_ablation} for details.}
    \label{fig:variants}
    \vspace{-1em}
\end{figure}

\section{Evaluation}
\label{sec:eval}

\begin{figure*}[t]
    \centering
    \begin{subfigure}[t]{0.32\textwidth}
        \centering
        \includegraphics[width=\linewidth]{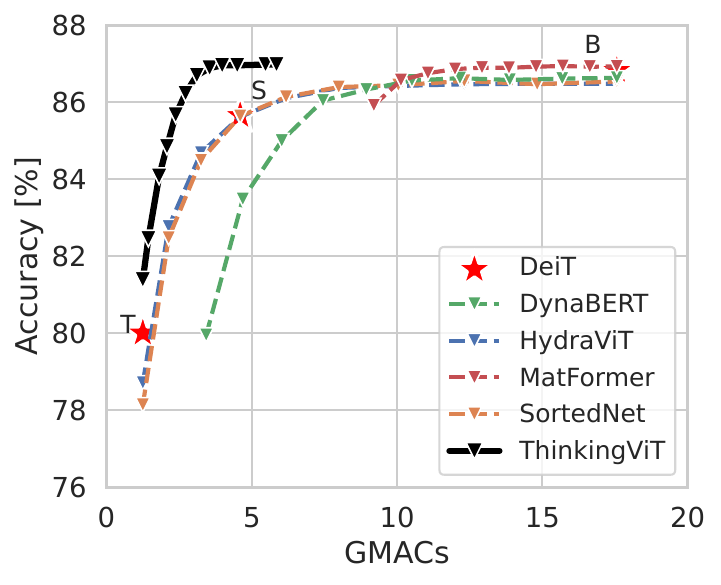}
        \caption{ImageNet-Real}
        \label{fig:real}
    \end{subfigure}
    \hfill
    \begin{subfigure}[t]{0.32\textwidth}
        \centering
        \includegraphics[width=\linewidth]{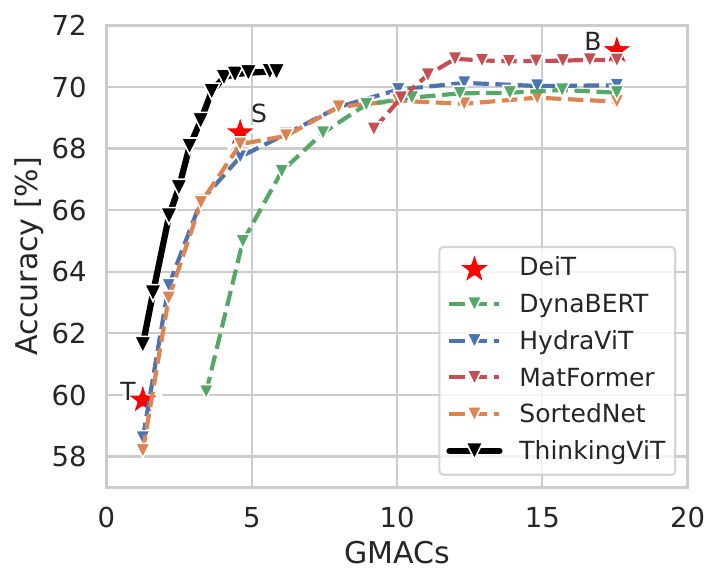}
        \caption{ImageNet-V2}
        \label{fig:v2}
    \end{subfigure}
    \hfill
    \begin{subfigure}[t]{0.32\textwidth}
        \centering
        \includegraphics[width=\linewidth]{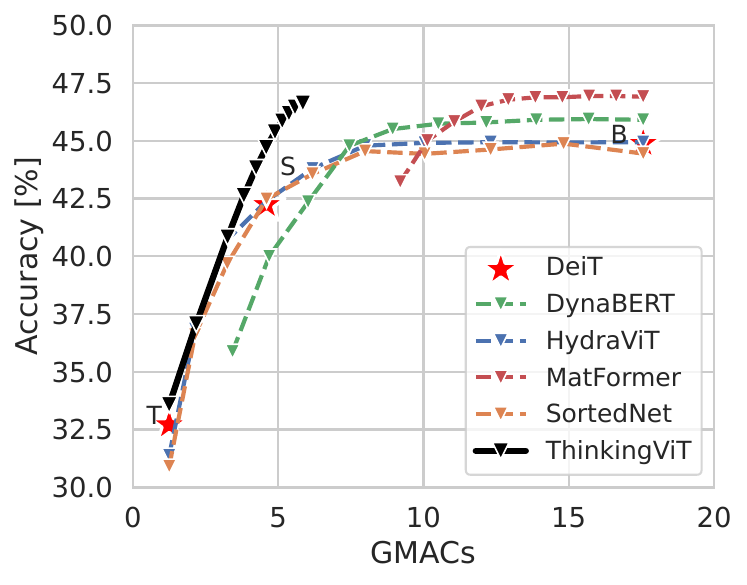}
        \caption{ImageNet-R}
        \label{fig:r}
    \end{subfigure}
    \caption{GMACs vs. Accuracy on ImageNet variants. \name has superior performance compared to the baselines.}
    \label{fig:gmacs_variants}
\end{figure*}

In this section, we analyze how entropy guides \name's adaptive computation (Sec.~\ref{sec:trade_offs}), evaluate it on ImageNet-1K and its variants against SOTA nested baselines (Sec.~\ref{sec:width_pruning}), extend the evaluation to segmentation architectures (Sec.~\ref{segmentation}) and hierarchical architectures such as Swin Transformer (Sec.~\ref{swin}), compare with early-exit baselines (Sec.~\ref{sec:early_exit_comparison}), and finally assess robustness on uniformly challenging datasets (Sec.~\ref{hard_samples}). 
 
\subsection{Setup}
\textbf{Implementation details:}
We train \name on ImageNet-1K \citep{ILSVRC15} at a resolution of $224 \times 224$, using models implemented in \texttt{timm} \citep{rw2019timm} and following the setup of \citet{touvron2021training}, with all models initialized with a pretrained DeiT-Tiny checkpoint. We train on a cluster with NVIDIA H100 GPUs, taking about 10 minutes per epoch on 2 GPUs.

\textbf{Baselines:} We compare \name against several width expansion baselines: MatFormer~\citep{devvrit2024matformer}, which slices only the hidden layer of the \gls{MLP} while keeping \gls{MHA} and embedding dimensions fixed;
DynaBERT~\citep{hou2020dynabert}, which slices both \gls{MHA} and the hidden layer of \gls{MLP} while keeping embeddings fixed;
SortedNet~\citep{valipour2023sortednet}, which slices all embeddings,  \gls{NORM}, \gls{MHA}, and \gls{MLP}, while keeping the number of heads fixed;
and HydraViT~\citep{haberer2024hydravit}, which slices \gls{MHA}, embeddings, \gls{NORM}, and \gls{MLP}, while also changing the number of heads.
Similar to HydraViT, \name adopts a slicing strategy across the number of heads, \gls{MHA}, embeddings, \gls{NORM}, and \gls{MLP} layers. However, unlike the above methods that follow static inference paths regardless of input difficulty, \name introduces input-adaptive computation.
In Section~\ref{sec:early_exit_comparison}, we present comparisons to depth-expansion baselines.

\textbf{Throughput and GMACs:}
We measure throughput and computational cost on a single NVIDIA A100 GPU with a batch size of 512.
To evaluate throughput, we run the model under different confidence thresholds~$\tau$. 
For each threshold, we record the per-image inference time across the entire validation set and compute the average throughput as the total number of images processed divided by the total inference time. 
By varying the threshold~$\tau$, we obtain the throughput curves reported in the plots.
For further details, see Appendix~\ref{sec:appendix_throughput}.
For GMACs, we follow a similar procedure.
The total compute per image for threshold~$\tau$ is then averaged over all samples as
\[
\text{GMACs}_{\tau} = \sum_{i=1}^{N} p_i \cdot \left( \sum_{j=1}^{i} \text{GMACs}_{j} \right),
\]
where $p_i$ denotes the fraction of samples that stop at round~$i$, and $\text{GMACs}_{j}$ denotes the GMACs of round $j$. 
By varying the threshold~$\tau$, we obtain the GMACs curves reported in the plots.

\begin{figure}[t]
  \centering
   \includegraphics[trim=10 10 10 0, clip, width=.45\textwidth]{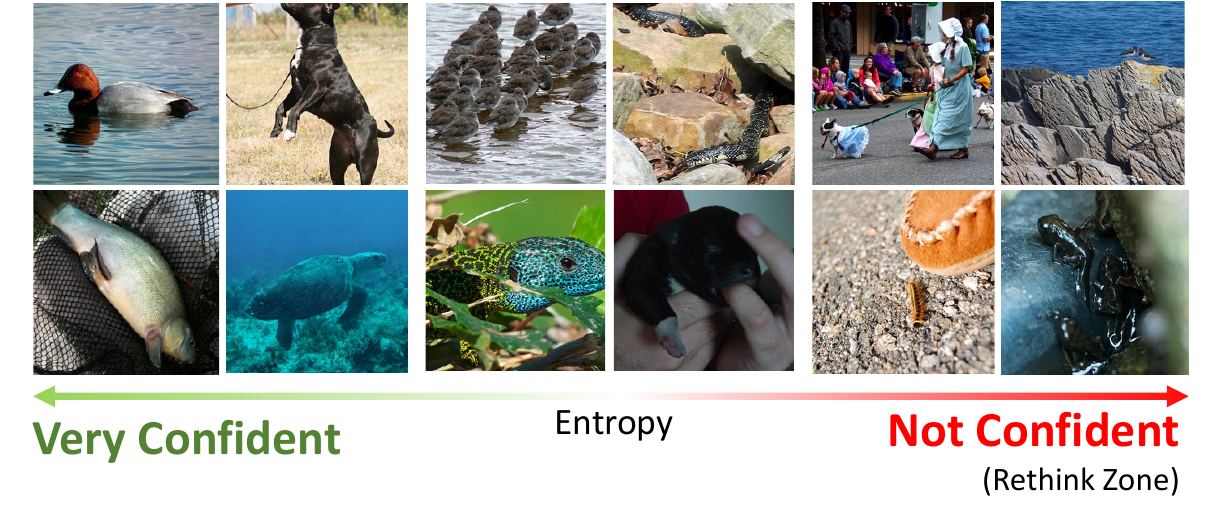}
  \caption{Visualization of images sorted by first-round entropy. \name confidently classifies simple, clear images in one round, while complex cases with occlusion or clutter show higher entropy and trigger a second round.}
   \label{fig:entropy_example}
   
   \vspace{-1em}
\end{figure}

\subsection{Trade-offs in attention expansion}
\label{sec:trade_offs}

\name expands computational capacity through progressive activation of attention heads. 
Generally, the number of thinking stages and the attention head expansion step size are hyperparameters, and their optimal configuration depends on the dataset and target efficiency objectives.
In Figure~\ref{fig:variants}, we evaluate this trade-off and find that the 3H~$\rightarrow$~6H configuration offers the most favorable balance between accuracy and compute on ImageNet-1K. The 2H~$\rightarrow$~3H~$\rightarrow$~6H variant spans the widest GMAC range among the high-performing models, while incurring only a small drop in accuracy compared to 3H~$\rightarrow$~6H. At the upper end, 3H~$\rightarrow$~6H~$\rightarrow$~12H achieves the highest final accuracy, surpassing 3H~$\rightarrow$~6H by 0.91 percentage points (p.p.), but raises the compute cost from 5.85 GMACs to 23.41 GMACs for only a marginal gain.
This substantial increase in cost leads to lower overall efficiency compared to the 3H~$\rightarrow$~6H setup, underscoring the importance of aligning \nameNoSpace’s thinking strategy with specific deployment goals.
Since we focus on configurations that offer the best trade-off between accuracy and GMACs, we adopt 3H~$\rightarrow$~6H for subsequent experiments.
For additional experiments and a larger depiction of Figure~\ref{fig:variants}, see Appendix~\ref{sec:thinking_ablation}.

\begin{figure*}[t]
  \centering
   \includegraphics[trim=2 0 0 0, clip, width=.8\textwidth]{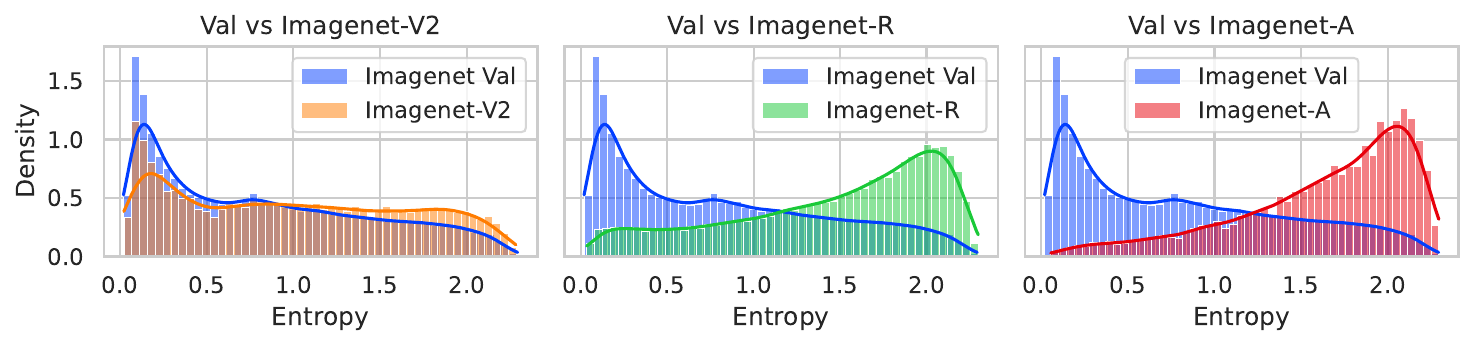}
  \caption{Entropy distribution after the first inference round across ImageNet validation sets. Simpler datasets like ImageNet-V2 show confident early predictions (left-skewed), while harder ones like ImageNet-A and -R show greater uncertainty (right-skewed), which triggers \name to go for the next round of thinking.}
   \label{fig:entropy_distribution}
   \vspace{-1.5em}
\end{figure*}

\begin{figure}[t]
    \centering
    \includegraphics[width=0.9\linewidth]{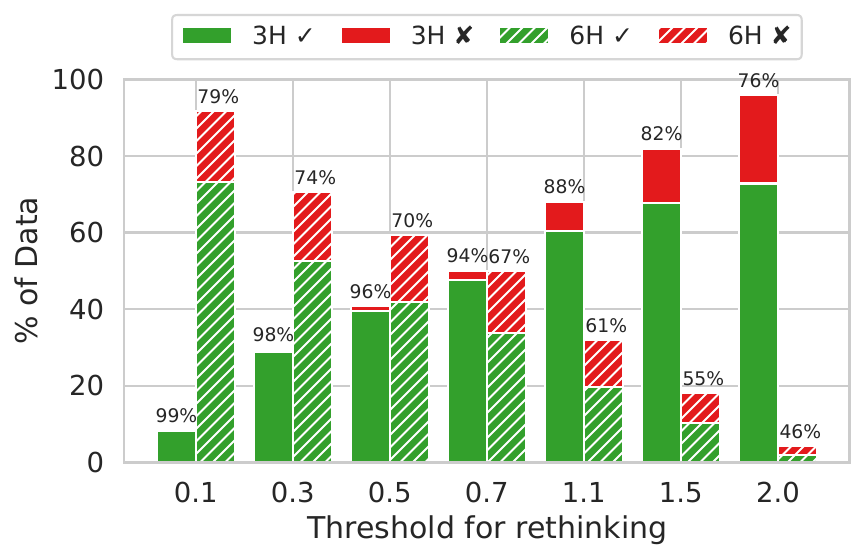}
    \caption{Load distribution of \name 3H~$\rightarrow$~6H across entropy thresholds. 
    Bars indicate load distribution; numbers show accuracy (green: correct, red: incorrect) on ImageNet-1K. 
    High 3H accuracy at low entropy confirms the effectiveness of entropy-based routing.}
    \label{fig:load}
\end{figure}

\begin{figure}[t]
    \centering
    \begin{subfigure}[t]{0.23\textwidth}
        \centering
        \includegraphics[width=\linewidth]{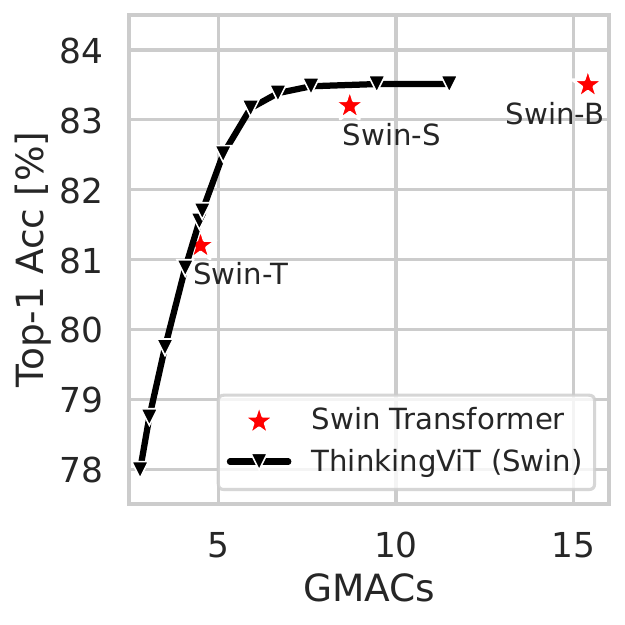}
        \caption{Swin Transformer}
        \label{fig:swin}
    \end{subfigure}
    \hfill
    \begin{subfigure}[t]{0.23\textwidth}
        \centering
        \includegraphics[width=\linewidth]{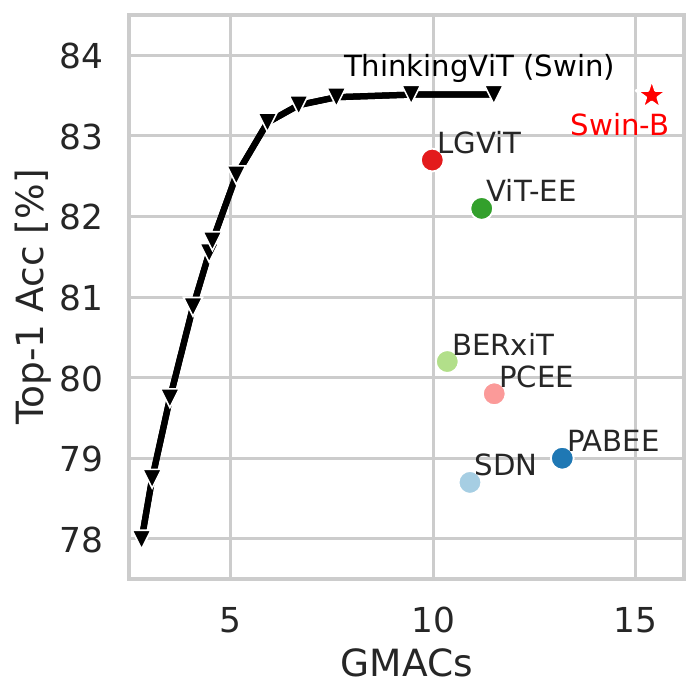}
        \caption{Early-Exit}
        \label{fig:early_exit}
    \end{subfigure}
    \caption{(a): \name with a Swin backbone outperforms standard Swin models, 
    matching Swin-Base accuracy with far fewer GMACs. 
    (b): Comparison of \name with Early-Exit baselines (with Swin backbones), 
    showing that iterative expansion provides improvement beyond early termination strategies.}
    \label{fig:swin_dynamicvit}
\end{figure}

\begin{figure*}[t]
    \centering
    \begin{subfigure}[t]{0.24\textwidth}
        \centering
        \includegraphics[width=\linewidth]{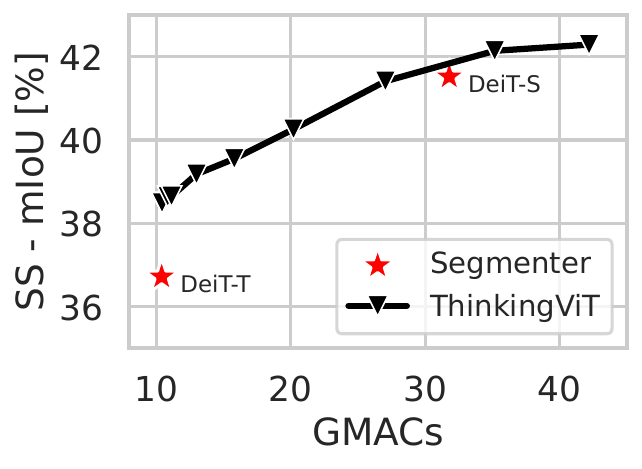}
        \caption{ADE20K (single-scale)}
        \label{fig:sub1}
    \end{subfigure}
    \hfill
    \begin{subfigure}[t]{0.24\textwidth}
        \centering
        \includegraphics[width=\linewidth]{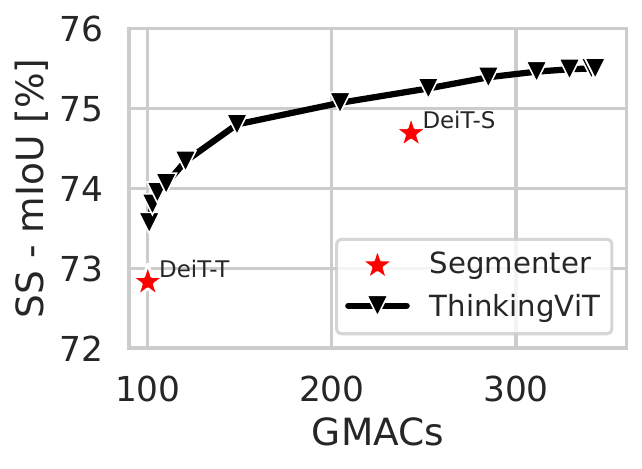}
        \caption{Cityscapes (single-scale)}
        \label{fig:sub2}
    \end{subfigure}
    \hfill
    \begin{subfigure}[t]{0.24\textwidth}
        \centering
        \includegraphics[width=\linewidth]{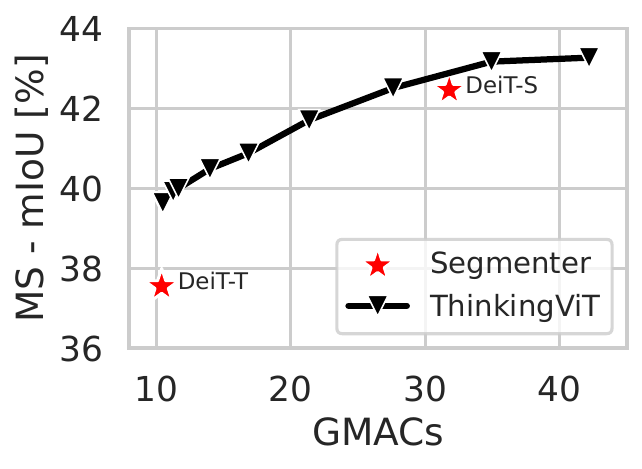}
        \caption{ADE20K (multi-scale)}
        \label{fig:sub3}
    \end{subfigure}
    \hfill
    \begin{subfigure}[t]{0.24\textwidth}
        \centering
        \includegraphics[width=\linewidth]{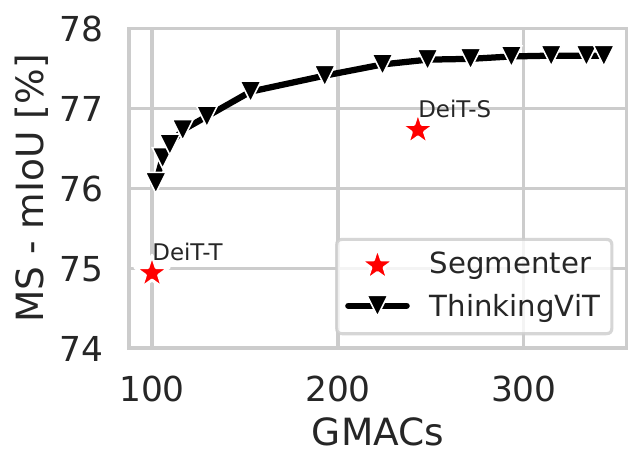}
        \caption{Cityscapes (multi-scale)}
        \label{fig:sub4}
    \end{subfigure}
\caption{
\textbf{\name performance on semantic segmentation.}
Comparison of mIoU versus GMACs on the ADE20K~\cite{zhou2017scene} and Cityscapes~\cite{cordts2016cityscapes} datasets under single- and multi-scale evaluation using the Segmenter~\cite{strudel2021segmenter} pipeline.
Similar to classification, \name achieves a favorable trade-off between GMACs and accuracy, and outperforms the DeiT-based backbones.
}
    \label{fig:semantic_segmentation}
\end{figure*}

\subsection{Comparing \name and baselines}
\label{sec:width_pruning}
Figure~\ref{fig:gmacs_throughput_val} compares \name with SOTA baselines in terms of GMACs and inference throughput, using the 3H~$\rightarrow$~6H configuration.
\name achieves up to 2.0~p.p.~higher accuracy at equal throughput, and up to 2.9~p.p.~higher accuracy at the same GMACs compared with its baselines. This improvement stems from three key factors.
\emph{First}, \name makes early predictions using only 3 heads for easy inputs and expands to 6 heads only when needed.
\emph{Second}, \name fuses embeddings from the first stage into the second, leading to better performance than flat 6-head models like HydraViT or DeiT-S.
Moreover, \name achieves 81.44\% with only 22.01 M parameters, which is just 0.36 p.p. lower than DeiT-Base with 86.6 M parameters. This result further validates the effectiveness of the Token Recycling strategy. Importantly, it also demonstrates that achieving higher accuracy does not necessarily require a larger model; rather, it depends on applying sufficient computation. 
\emph{Third}, baselines typically employ multiple slicing points throughout the architecture to meet diverse GMACs. However, optimizing for all such configurations can lead to accuracy reduction \citep{haberer2024hydravit}. \name sidesteps this issue by supporting elastic inference using just a few nested subnetworks, with compute budget adaptability controlled via the entropy threshold. We report detailed performance results in the Appendix~\ref{appendix_detailed_results}.

To assess robustness, we evaluate \name on ImageNet-V2~\citep{recht2019imagenet}, ImageNet-ReaL~\citep{beyer2020imagenet}, and on ImageNet-R~\citep{hendrycks2020many} in Figure~\ref{fig:gmacs_variants}. Additional results on ImageNet-A~\citep{hendrycks2019nae} and ImageNet-Sketch~\citep{wang2019learning} are provided in Appendix~\ref{appendix_variants}.
Across all robustness benchmarks, \name consistently surpasses the baselines, demonstrating superior generalization under distribution shifts.
Furthermore, on ImageNet-ReaL, -Sketch, and -R, \name exceeds the accuracy of DeiT-Base, despite using significantly fewer parameters (22.1M vs.\ 86.6M) and lower GMACs (5.85 vs.\ 17.56), which highlights the effectiveness of the \emph{Token Recycling}. 
We further compare and combine \name with "token-based pruning" approaches in Appendix~\ref{appendix:token_pruning_baseliens}, showing that \name both outperforms these methods and remains complementary with them.


\subsection{Analyzing entropy as a signal to think deeper}
After each stage, \name uses output entropy to assess certainty and decide if further computation is needed.
Figure~\ref{fig:entropy_distribution} shows the entropy after the first round of inference on ImageNet-A~\citep{hendrycks2019nae}, ImageNet-V2~\citep{recht2019imagenet}, and ImageNet-R~\citep{hendrycks2020many}, providing insight into how \name 3H~$\rightarrow$~6H estimates uncertainty across different distribution shifts.
On easier datasets like ImageNet-V2, entropy is left-skewed, reflecting confident early predictions. In contrast, harder datasets like ImageNet-A and -R produce right-skewed distributions, triggering deeper inference in \nameNoSpace.
In Figure~\ref{fig:load}, we show the load distribution on the 3H and 6H submodels under varying entropy thresholds. As entropy decreases, fewer samples proceed to the second stage, and those that exit early are rarely misclassified. This confirms that entropy effectively halts computation on easy inputs without sacrificing accuracy.
Furthermore, Figure~\ref{fig:entropy_example} shows images with different entropy levels to illustrate how \name adjusts its computation. Easy images such as clear pictures of a single object have low entropy and stop after one round of thinking. More difficult images, including those with several objects, blocked views, or poor lighting, show higher entropy and trigger a second round of thinking to improve accuracy.


\subsection{Semantic Segmentation}
\label{segmentation}
\begin{figure}[t]
    \centering
    \includegraphics[width=.75\linewidth]{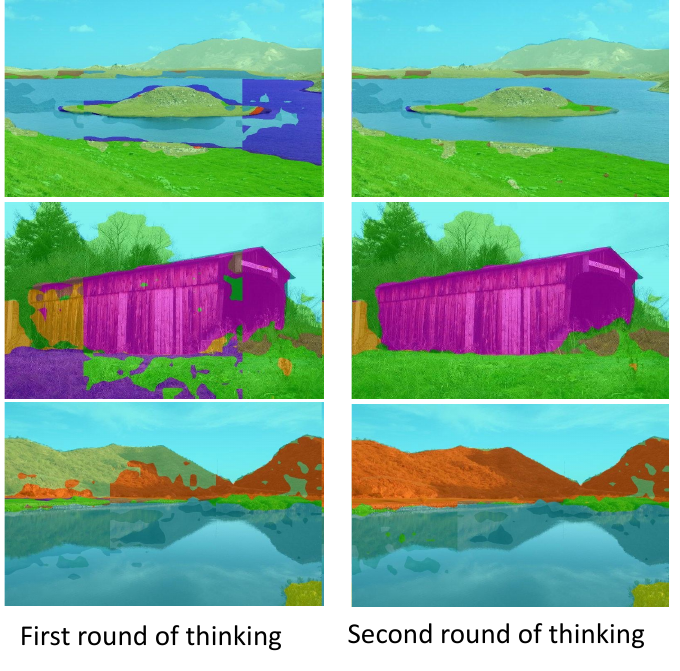}
\caption{
\textbf{ThinkingViT segmentation after one and two rounds of thinking.}
Example outputs from the ADE20K \cite{zhou2017scene} dataset. The second round refines object boundaries and improves segmentation quality compared to the first round.
}
    \label{fig:segm}
    \vspace{-1em}
\end{figure}

\name builds on the vanilla ViT architecture and is therefore compatible with models designed for ViT backbones. To verify this, we replace the ViT backbone in the \textit{Segmenter} \cite{strudel2021segmenter} model with \name and employ a linear decoder for semantic segmentation. We evaluate on the ADE20K \cite{zhou2017scene} and Cityscapes \cite{cordts2016cityscapes} datasets using both single-scale and multi-scale inference. As shown in Fig.~\ref{fig:semantic_segmentation}, \name consistently outperforms DeiT-Small and DeiT-Tiny backbones, demonstrating its effectiveness as a drop-in replacement for vision transformer architectures in downstream tasks. 
Furthermore, Fig.~\ref{fig:segm} visualizes how \name refines object boundaries and improves segmentation accuracy with additional rounds of "thinking". \name is also compatible with \gls{ViT} based tasks such as DETR~\cite{carion2020end} style object detection, which we leave for future work. 


\subsection{Extension to Hierarchical Architectures}
\label{swin}


While all previous experiments were based on a vanilla ViT architecture, we also demonstrate that \name can be applied to more modern variants of Vision Transformers, such as the Swin Transformer~\cite{liu2021swin}. The key difference in Swin Transformer is that it consists of four stages, each with an increasing number of attention heads and a decreasing number of tokens, which grow progressively larger. We integrate \name on top of Swin-S, by allocating the entire first stage and half of the capacity of the last three stages for the first round, as the latter contains most of the layers and attention heads. The second round is then executed using the full model. As illustrated in Figure~\ref{fig:swin}, this design allows \name with a Swin backbone to achieve the same accuracy as Swin-B (83.5\%) while requiring only 9.5~GMACs and 50M parameters, compared to 15.4~GMACs and 88M parameters for Swin-B. 
Moreover, our model achieves a higher accuracy per GMAC compared to both Swin-S and Swin-T, demonstrating that our approach can improve performance even for hierarchical architectures.

\subsection{Comparison with Early-Exit Models}
\label{sec:early_exit_comparison}

\name reduces computation by adapting model \textit{width}, so its primary baselines are width-pruning approaches (Section~\ref{sec:width_pruning}). 
To additionally compare against depth-adaptive approaches, we evaluate \name against several early-exit models, including SDN~\citep{kaya2019shallow}, PABEE~\citep{zhou2020bert}, BERxiT~\citep{xin2021berxit}, ViT-EE~\citep{bakhtiarnia2021multi}, PCEE~\citep{zhang2022pcee}, and LGViT \cite{xu2023lgvit}. As shown in Fig.~\ref{fig:early_exit}, \name achieves much higher accuracy across comparable GMAC budgets. 
In Appendix~\ref{appendix_Comparison_with_Early_Exit}, we provide a detailed comparison with BranchyNet \cite{teerapittayanon2016branchynet}.

\subsection{\name performance on hard inputs}
\label{hard_samples}

When the input distribution is uniformly challenging, most samples require deeper inference. To evaluate \name under such conditions, we use ImageNet-R, a benchmark constructed from examples "misclassified" by ResNet-50~\citep{he2016deep}. Since ResNet-50 has an accuracy similar to the 3H subnetwork, these samples serve as hard cases for the first stage of \name and can simulate a uniformly hard input setting. 
Figure~\ref{fig:entropy_distribution} shows that \name routes most samples to the expanded 6H stage in this case. 
This demonstrates that \nameNoSpace’s entropy-based routing behaves as intended. Additionally, Figure~\ref{fig:r} shows that \name still outperforms all of the baselines in these hard conditions.

\section{Conclusion}
\label{sec:conclusion}
We introduce \nameNoSpace, a nested \gls{ViT} that uses progressive thinking stages to adapt inference computation to input difficulty. The model begins with a small subset of important attention heads and halts early when confidence is high. Otherwise, within the same backbone, it activates a larger subset of attention heads and conducts a new forward pass. This iterative expansion continues until the model reaches the desired confidence level or exhausts its maximum capacity. Token Recycling further improves later stages by fusing input embeddings with features from previous stages. \name outperforms both width-based and depth-based expansion strategies, and its backbone-preserving design enables seamless transfer to downstream tasks such as semantic segmentation. Experiments show that \name surpasses nested baselines by up to 2.0 percentage points (p.p.) in accuracy at the same throughput and by up to 2.9 p.p. at equal GMACs on ImageNet-1K. We also demonstrate that it extends effectively to architectures such as Swin Transformer.


\section*{Acknowledgements}

This research received funding from the Federal Ministry for Digital and Transport under the CAPTN-Förde 5G project (grant no.~45FGU139H), the German Ministry of Transport and Digital Infrastructure through the CAPTN Förde Areal II project (grant no.~45DTWV08D), Federal Ministry for Economic Affairs and Energy under the CAPTN X-FERRY project (grant no.~FK: 03SX612A), and the Federal Ministry for Economic Affairs and Climate Action under the Marispace-X project (grant no.~68GX21002E). It was supported in part by high-performance computing resources provided by the Kiel University Computing Centre and the Hydra computing cluster, funded by the German Research Foundation (grant no.~442268015) and the Petersen Foundation (grant no.~602157).

{
    \small
    \bibliographystyle{ieeenat_fullname}
    \bibliography{main}
}

\clearpage
\setcounter{page}{1}
\maketitlesupplementary
\appendix

\begin{table*}[t]
\centering
\small
\caption{Comparison of \name and BranchyNet on DeiT.}
\label{tab:early_exit_comparison}
\begin{tabular}{lcccc}
\toprule
\textbf{Model} & \textbf{GMACs} & \textbf{Throughput [\#]} & \textbf{Params [M]} & \textbf{Accuracy [\%]}\\

\midrule
DeiT-Tiny & 1.25 & 10047.6 & 5.7 & 72.2 \\
DeiT-Small & 4.6 & 4603.6 & 22.1 & 79.9 \\
\midrule
BranchyNet 3H & 1.25 & 10047.6 & 27.8 & 74.51 \\
BranchyNet 3H~$\rightarrow$~6H & 5.85 & 3157.1 & 27.8 & 78.08 \\
\midrule
\textbf{\name 3H}    & 1.25 & 10047.6 & 22.1 & 73.58  \\
\textbf{\name 3H~$\rightarrow$~6H}    & 5.85 & 3157.1 & 22.1 &  \textbf{81.44} \\
\bottomrule
\end{tabular}
\end{table*}

\section{Adaptive inference performance of \name under varying entropy thresholds}
\label{appendix_detailed_results}

Table~\ref{tab:detailed_results} presents detailed results of \nameNoSpace’s adaptive inference behavior under different entropy thresholds. We report accuracy, number of parameters, throughput, total compute (in GMACs), and the ratio of inputs that proceed to the second stage of inference of \name 3H~$\rightarrow$~6H. 

\begin{table*}
\centering
\small
\caption{Performance metrics of \name 3H~$\rightarrow$~6H across different entropy thresholds}
\begin{tabular}{cccccc}
\toprule
\textbf{Entropy} & 
\multirow{2}{*}{\textbf{Accuracy}} & 
\textbf{Throughput} & 
\textbf{Params} & 
\multirow{2}{*}{\textbf{GMACs}} & 
\textbf{Second Round} \\
\textbf{Threshold} &  & \textbf{[\#/s]} & \textbf{[M]} &  & \textbf{Call Ratio [\%]} \\
\midrule
0     & 81.444 & 3157.09  & 22.01 & 5.85 & 100.0 \\
0.1   & 81.440 & 3347.69  & 22.01  & 5.47   & 91.7  \\
0.3   & 81.438 & 3955.05  & 22.01  & 4.50  & 70.58  \\
0.5   & 81.386 & 4380.71  & 22.01  & 3.98  & 59.29  \\
0.7   & 81.230 & 4807.04  & 22.01  & 3.55  & 49.95  \\
0.9   & 80.714 & 5342.47  & 22.01  & 3.11  & 40.36  \\
1.1   & 79.990 & 5918.90  & 22.01  & 2.72  & 31.97  \\
1.3   & 79.114 & 6535.13  & 22.01  & 2.38 & 24.63  \\
1.5   & 77.936 & 7201.46  & 22.01   & 2.08  & 18.11  \\
1.7   & 76.766 & 7944.38  & 22.01   & 1.81  & 12.13  \\
2     & 74.736 & 9203.90  & 22.01  & 1.44  & 4.20   \\
2.5   & 73.580 & 10047.60 & 22.01  & 1.25  & 0.0  \\
\bottomrule
\end{tabular}
\label{tab:detailed_results}
\end{table*}

\section{Results on ImageNet variants}
\label{appendix_variants}
\begin{figure}[h]
    \centering
    \includegraphics[width=0.45\textwidth]{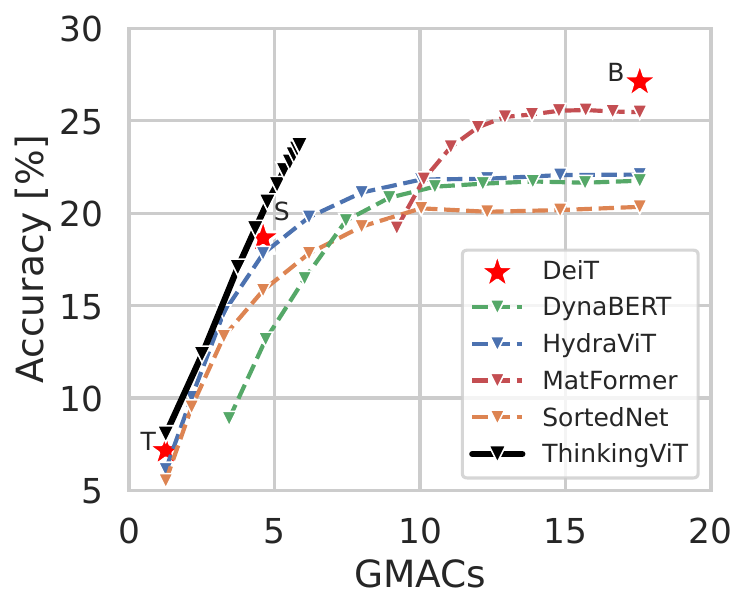}
    \caption{Full results of \name and baselines in terms of GMACs on ImageNet-A.} 
    \label{fig:a}
\end{figure}

\begin{figure}[h]
    \centering
    \includegraphics[width=0.45\textwidth]{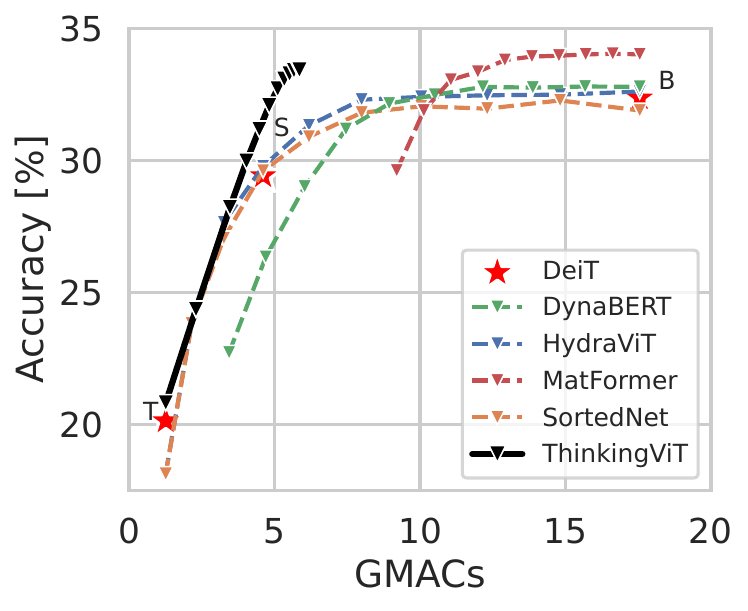}
    \caption{Full results of \name and baselines in terms of GMACs on ImageNet-Sketch.} 
    \label{fig:sketch}
\end{figure}

Figure~\ref{fig:a} and ~\ref{fig:sketch} show the performance of \name on ImageNet variants. Note that the GMACs for ImageNet-V2 and ImageNet-R are reported in Figure~\ref{fig:v2} and Figure~\ref{fig:r}, respectively.

\section{Comparison with BranchyNet}
\label{appendix_Comparison_with_Early_Exit}

Table~\ref{tab:early_exit_comparison} compares \name with BranchyNet, a standard early exit model~\citep{teerapittayanon2016branchynet}, applied to DeiT \citep{touvron2022deit} that matches its GMACs and throughput. The baseline consists of 24 layers: the first 12 use 3 attention heads (3H) and the remaining 12 use 6 heads (6H). Similar to \name, the exits are jointly trained together. At the 3H point, \name records slightly lower accuracy because the first three heads must generate representations that remain compatible with the later expansion to six heads. Additionally, their weights must be trained in a way that allows the weights of the subsequent three heads to build upon them.

After expanding to 6 attention heads, \name reaches 81.44\% top-1 accuracy, improving by 3.37 p.p. upon the early exit baseline, which attains 78.08\%. This improvement likely stems from \name increasing the number of active attention heads within a fixed 12-layer backbone, allowing gradients to pass through fewer layers (12 layers compared to 24 layers) and reducing some of the optimization challenges associated with deeper networks. These findings indicate that allocating compute along the width dimension by gradually increasing attention heads can be more effective than depth-based early exit strategies used in prior work. 
Further, since it adjusts model width rather than depth, \name is complementary to early-exit strategies and can be combined with them to further expand deployment flexibility.



\section{Ablation of different token recycling strategies}
\label{appendix_token_recycling_ablation}

In Table~\ref{tab:thinking_designs}, we compare several design variants for recycling information from the first round of inference to get better performance in the second round. We conduct these experiments on \name 3H~$\rightarrow$~6H. In the \textit{Layerwise Activation Snapshots} variant, we cache the hidden states at each of the 12 layers from the first round and feed them into the corresponding layers in the second round. 
However, this approach performs suboptimally, likely because it forces all intermediate representations from the first round to be reused in the second round, which makes it harder for the tokens as they need to work for both rounds.
In the \textit{Memory Tokens} design, we introduce a set of learnable memory tokens \citep{darcet2024vision} that, in the first round, store helpful information for the second round.
We also experiment with hybrid strategies that combine activation snapshots with Memory Tokens or introduce a fresh \texttt{[CLS]} token in the second round.
We additionally evaluated the use of KV-cache reuse \cite{vaswani2017attention}, originally introduced for generative models, and include the results in Table~\ref{tab:thinking_designs}.
Ultimately, we find that the simplest strategy, \textit{Final-Layer Token Recycling}, which reuses the features from the last layer of the first round, on average, achieves the best performance.
This indicates that the output of the first round already captures sufficient high-level information to guide the second round effectively, while being comparatively simple to implement. It is important to note that except the KV-cache experiment, all results were conducted during the early experimental phase of \name, and due to training resource constraints, they were not trained with the full joint training strategy described in Section~\ref{training_jointly}, but rather with the stochastic training method introduced in~\cite{haberer2024hydravit}. As a result, their accuracies are slightly lower than those of the final model reported in Section \ref{sec:eval}.

\begin{table*}
\centering
\small
\caption{Comparison of design variants for conditioning the second round of inference on the first.}
\vspace{0.4em}
\begin{tabular}{lcc}
\toprule
\multirow{2}{*}{\textbf{Design Variant}} & 
\textbf{First Thought} & 
\textbf{Second Thought} \\
 & \textbf{Acc. [\%]} & \textbf{Acc. [\%]} \\
\midrule
    \textit{DeiT Baseline (Tiny → Small)} & 72.2 & 79.9 \\
\midrule
    Layerwise Activation Snapshots & 73.794 & 78.566 \\ 
    Memory Tokens & 73.96 & 78.9 \\ 
    Layerwise Activation Snapshots + new \texttt{[CLS]} in second round & 73.58 & 77.94 \\ 
    Layerwise Activation Snapshots + Memory Tokens & 73.71 & 79.04 \\ 
    KV-Caching \cite{vaswani2017attention} & 73.872 & 75.674 \\ 
    Final-Layer Token Recycling (\name) & 73.13 & \textbf{80.05} \\ 

    \Xhline{2\arrayrulewidth}
\end{tabular}
\label{tab:thinking_designs}
\end{table*}

\section{Ablation of different fusing strategies}
\label{appendix_fusing_ablation}

Table~\ref{tab:fusion_designs_accuracy} compares six fusion strategies that recycle the tokens produced in the first round, denoted as $z_{1}$, by incorporating them into the initial patch embeddings of the second round, $\mathcal{E}_{2}$, on \name with a configuration of $3H\rightarrow 6H$.
Each variant forms the second-round input as a linear blend, where $\alpha$ is a learnable scalar. The strategies differ in how the two embeddings are projected to the same dimensional space. We consider two projection strategies:
(I) a parameter-free approach that either repeats the embedding or pads the lower-dimensional embeddings with zeros, and
(II) a learnable linear projection. Empirically, the learnable projection achieves the best overall accuracy by having the highest accuracy in the second round while maintaining performance comparable to the strongest model configuration in the first round.
In addition, we evaluate initializing $\alpha = 1$ and observe that initializing with zero consistently leads to the best performance.
This initialization allows the model to begin training without relying on information from the first round, thereby enabling it to learn the optimal degree of reuse.
For example, in \name configured with 3H~$\rightarrow$~6H heads, the learned value of $\alpha$ converges to $-0.19$.

\begin{table*}
\centering
\small
\caption{Impact of different fusion strategies for integrating first-round tokens ($z_{1}$) into second-round embeddings ($\mathcal{E}_{2}$) during progressive inference on \name with 3H~$\rightarrow$~6H.}
\label{tab:fusion_designs_accuracy}
\begin{tabular}{@{}p{2cm}p{2.8cm}p{4.1cm}cc@{}}
\toprule
\multirow{2}{*}{\textbf{Fusion Method}} & 
\multirow{2}{*}{\textbf{Dim Alignment}} & 
\multirow{2}{*}{\textbf{Note}} & 
\multicolumn{2}{c}{\textbf{Acc. [\%]}} \\
 &  &  & \textbf{1\textsuperscript{st} Round} & \textbf{2\textsuperscript{nd} Round} \\
\midrule
$\alpha \cdot z_{1}+ \mathcal{E}_{2}$ & Pad $z_{1}$ with zeros &  Pad the first half, $init(\alpha)=0$ & 73.32\% & 81.37\% \\ 
$\alpha \cdot z_{1}+ \mathcal{E}_{2}$ & Pad $z_{1}$ with zeros & Pad the second half, $init(\alpha)=0$ & \textbf{74.12\%} & 80.32\% \\ 
$\alpha \cdot z_{1} + \mathcal{E}_{2}$ & Repeat $z_{1}$ & $init(\alpha)=1$ & 72.99\% & 80.87\% \\ 
$\alpha \cdot z_{1}+ \mathcal{E}_{2}$ & Repeat ${z_{1}}$  & $init(\alpha)=0$ & 73.14\% & 81.41\% \\ 
$z_{1}+ \alpha \cdot \mathcal{E}_{2}$& Repeat ${z_{1}}$ & $init(\alpha)=0$ & 72.89\% & 80.51\% \\
$\alpha \cdot z_{1}+ \mathcal{E}_{2}$ & Linear (\name) & $init(\alpha)=0$  &73.58\% & \textbf{81.44\%} \\ 
\bottomrule
\end{tabular}
\end{table*}

\section{Prediction dynamics across two inference rounds on ImageNet-1K}
\label{appendix_Qualitative_Analysis_Across_Inference_Rounds}

\begin{figure}[ht]
  \centering
   \includegraphics[trim=2 0 0 0, clip, width=.42\textwidth]{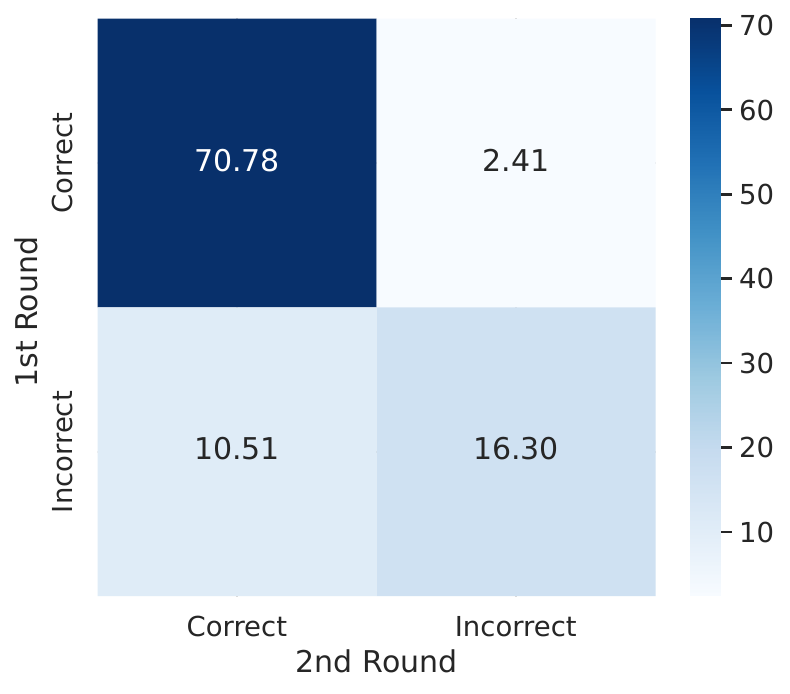}
  \caption{Prediction dynamics of \name 3H~$\rightarrow$~6H across two inference rounds on ImageNet-1K.}
   \label{fig:conf_matrix}
\end{figure}

Figure~\ref{fig:conf_matrix} presents prediction dynamics across two inference rounds of  \name 3H~$\rightarrow$~6H on the ImageNet-1K validation set. A small portion of samples, approximately 2\%, were correctly classified in the first round but misclassified in the second. This phenomenon, often attributed to \emph{overthinking}~\citep{kumar2025overthink}, is also observed in other architectures such as ResNet and Swin Transformers~\citep{ding2025occam}, where smaller models sometimes outperform larger ones on few samples by avoiding unnecessary complexity.
The majority of samples, around 70\%, were correctly classified in both rounds. These generally correspond to visually simple or unambiguous cases where one round of inference is sufficient. Roughly 10\% of samples were initially misclassified but corrected in the second round, demonstrating the benefit of additional reasoning for harder examples. Finally, around 16\% remained misclassified across both rounds, indicating that these inputs are intrinsically ambiguous or fall outside the model's capacity, even with increased computation.

\section{Comparing and combining \name with other adaptive baselines}
\label{appendix:token_pruning_baseliens}
\begin{figure}[h!]
        \centering
        \includegraphics[width=0.45\textwidth]{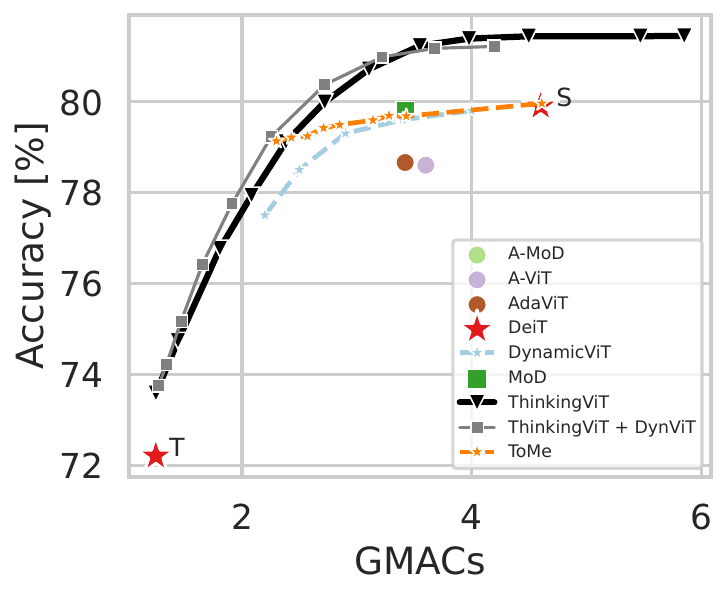}
        \caption{Comparing and combining \name with token pruning baselines.}
        \label{fig:gmacs_tp_baselines}
\end{figure}

\name\ builds upon a nested backbone to enable input adaptivity, so our evaluation in Section~\ref{sec:eval} focuses on nested Transformer baselines.
For completeness, we also evaluate against several representative token-level dynamic models, including MoD~\citep{raposo2024mixture}, AMoD~\citep{gadhikar2025attention}, ToMe~\citep{bolya2023token}, A-ViT~\citep{yin2022vit}, AdaViT~\citep{meng2022adavit}, and DynamicViT~\citep{rao2021dynamicvit}. As shown in Figure~\ref{fig:gmacs_tp_baselines}, \name achieves greater scalability and consistently better GMACs-Accuracy tradeoffs.

We note that  \name performs routing at the \textit{image level}, which is orthogonal to token-level pruning \citep{han2023dynamic, wang2021not, wang2020glance}, and thus can be combined with such approaches to provide additional flexibility.
To demonstrate this, we incorporate the token pruning mechanism of DynamicViT~\cite{rao2021dynamicvit} into the second round of \name 3H~$\rightarrow$~6H, applying a pruning ratio of 0.8. 
As shown in Fig.~\ref{fig:gmacs_tp_baselines}, integrating DynamicViT further improves both efficiency and accuracy, indicating that \name 3H~$\rightarrow$~6H is compatible with and complementary to existing token pruning approaches.


\section{Comparison of accuracy vs. GMACs for baselines based on DeiT-Small}
\label{appendix_gmacs_imagenet_val_for_small}

\begin{figure}[h!]
        \centering
        \includegraphics[width=0.45\textwidth]{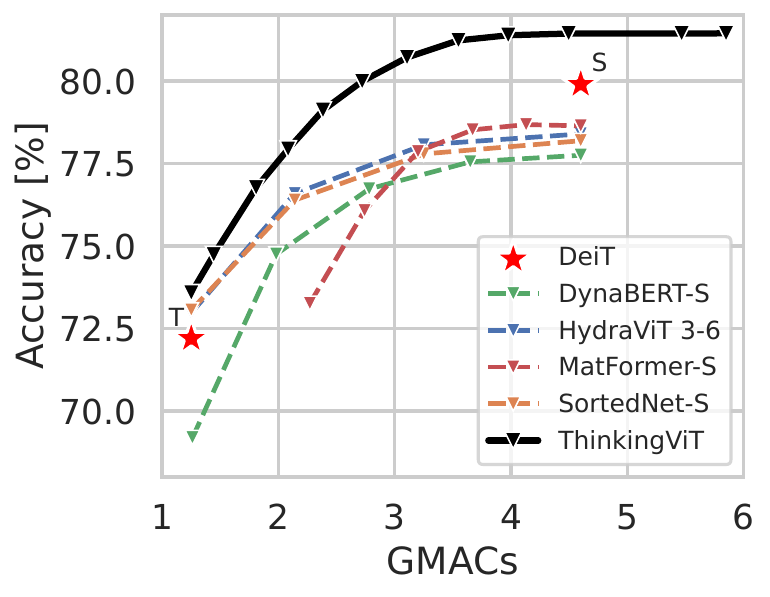}
        \caption{
Accuracy versus GMACs on the ImageNet-1K validation set. All baseline models are based on DeiT-Small. \name consistently outperforms these baselines by achieving higher accuracy at comparable or lower computational cost.
}
        \label{fig:gmacs_imagenet_val_for_small}
\end{figure}

In Figure~\ref{fig:gmacs_imagenet_val_for_small}, we compare accuracy versus GMACs on ImageNet-1K using DeiT-Small as the common backbone for all baselines. 
\name achieves higher accuracy at similar or lower compute budgets, showing a consistently more favorable tradeoff. This demonstrates that the benefits of \name hold across backbone scales.
\section{ImageNet-21K results}
\label{appendix_imagenet21k}

To evaluate \name on a larger-scale classification task, we train \name, MatFormer, HydraViT, DeiT-Tiny, and DeiT-Small on ImageNet-21K~\citep{ILSVRC15} as follows:

We use the shuffled dataset \texttt{imagenet-w21-wds}~\footnote{\url{https://huggingface.co/datasets/timm/imagenet-w21-wds}} with the first 12,741,248 images as the training split and the remaining 411,008 images as the validation split. All models are trained from scratch for 150 epochs with a global batch size of 2048.

\begin{figure}[h!]
        \centering
        \includegraphics[width=0.45\textwidth]{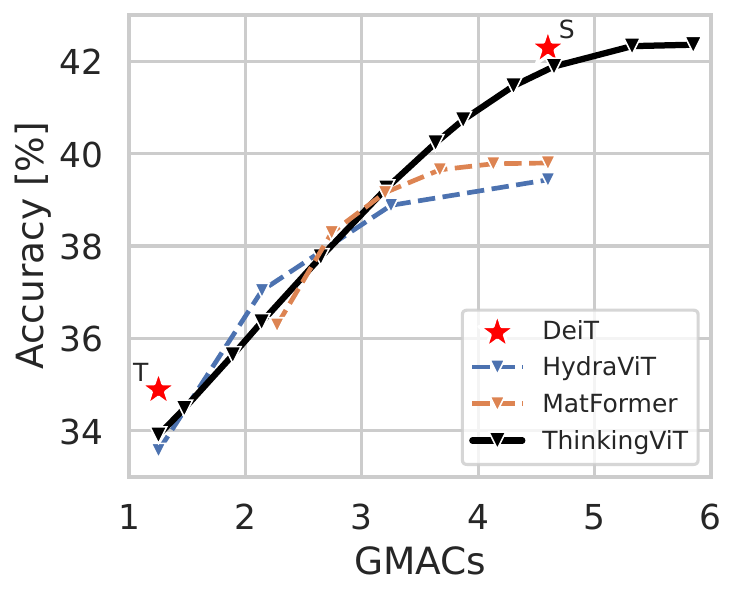}
        \caption{
Accuracy versus GMACs on ImageNet-21K. All models are trained from scratch for 150 epochs. While nested methods perform similarly below 3.5 GMACS, \name achieves a higher accuracy ceiling. Stars denote standalone DeiT-Tiny~(T) and DeiT-Small~(S).
}
        \label{fig:gmacs_plot_21k}
\end{figure}

Figure~\ref{fig:gmacs_plot_21k} reports accuracy versus GMACs on the validation split.
In the low-compute regime, i.e., below 3.5 GMACs, all nested methods perform comparably.
However, as the compute budget increases, \name pulls ahead: at roughly 4.6 GMACs, \name reaches 41.9\% accuracy, outperforming both MatFormer at 39.8\% and HydraViT at 39.4\% by more than 2 percentage points.
Moreover, \name offers a higher accuracy ceiling of 42.4\% by leveraging its second inference round for difficult samples, matching standalone DeiT-Small at 42.3\% while providing a continuous range of cheaper operating points through entropy-based early exit.

These results show that the benefits of \name generalize beyond ImageNet-1K to large-scale settings with an order of magnitude more classes.

\section{Limitations}
\label{sec:appendix_limitations}

\textbf{Training Overhead:} Compared to training multiple standalone models, nested models like \name incur higher training costs to reach similar performance levels across all stages. This is a known limitation of nested architectures, which share parameters and require joint optimization to maintain accuracy at varying compute levels. In our case, \name 3→6 trains DeiT-T and DeiT-S with a two-stage forward and single backward pass, taking ~20 min/epoch, comparable to training DeiT-T (7 min) and DeiT-S (13 min) separately. Therefore training \name takes as long as training DeiT-T and DeiT-S for 300 epochs each.

\noindent
\paragraph{Jumping Too Far Reduces Effectiveness:} When the model transitions from a very small subset of attention heads (e.g., 3H) directly to a full configuration (e.g., 12H) in the second stage, performance suffers compared to using an intermediate stage (e.g., 9H) instead. This suggests that overly aggressive compute expansion can disrupt representational continuity, emphasizing the importance of smooth progression in staged inference.

\section{Batch inference}
\label{sec:appendix_throughput}

After each “thinking” round, we filter out the samples whose entropy falls below the stopping threshold (i.e., confident predictions). The remaining uncertain samples are then re-batched and forwarded to the next round. Since all rounds share the same Transformer backbone and Token Recycling only adjusts the input embeddings, no additional batch scheduling is needed. As a result, \name integrates smoothly into standard batched inference engines, where the model proceeds in synchronized rounds and the batch size gradually decreases as samples exit early; see Algorithm~\ref{alg:batch_inference}.

\begin{algorithm}[h]
\caption{Batched Inference with \name}
\label{alg:batch_inference}
\begin{algorithmic}[1]
\State $\mathcal{B} \gets$ input batch
\For{$r = 1$ to $R$} \Comment{Progressive "thinking" rounds}
    \State $\hat{y} \gets \text{Forward}(\mathcal{B}, r)$
    \State $\mathcal{C} \gets \{ i \in \mathcal{B} \mid \text{Entropy}(\hat{y}_i) \le \tau \}$ \Comment{Confident samples}
    \State $\mathcal{U} \gets \mathcal{B} \setminus \mathcal{C}$ \Comment{Uncertain samples}
    \State \textbf{output} predictions for samples in $\mathcal{C}$
    \If{$\mathcal{U}$ is empty} \textbf{break}
    \EndIf
    \State $\mathcal{B} \gets$ Rebatch($\mathcal{U}$) \Comment{Shrink batch}
\EndFor
\end{algorithmic}
\end{algorithm}
\section{Effect of attention-head expansion and loss weighting on multi-round thinking}
\label{sec:thinking_ablation}

\begin{figure*}[t]
    \small
    \centering
    \begin{subfigure}{1\textwidth}
        \centering
        \includegraphics[width=0.9\textwidth]{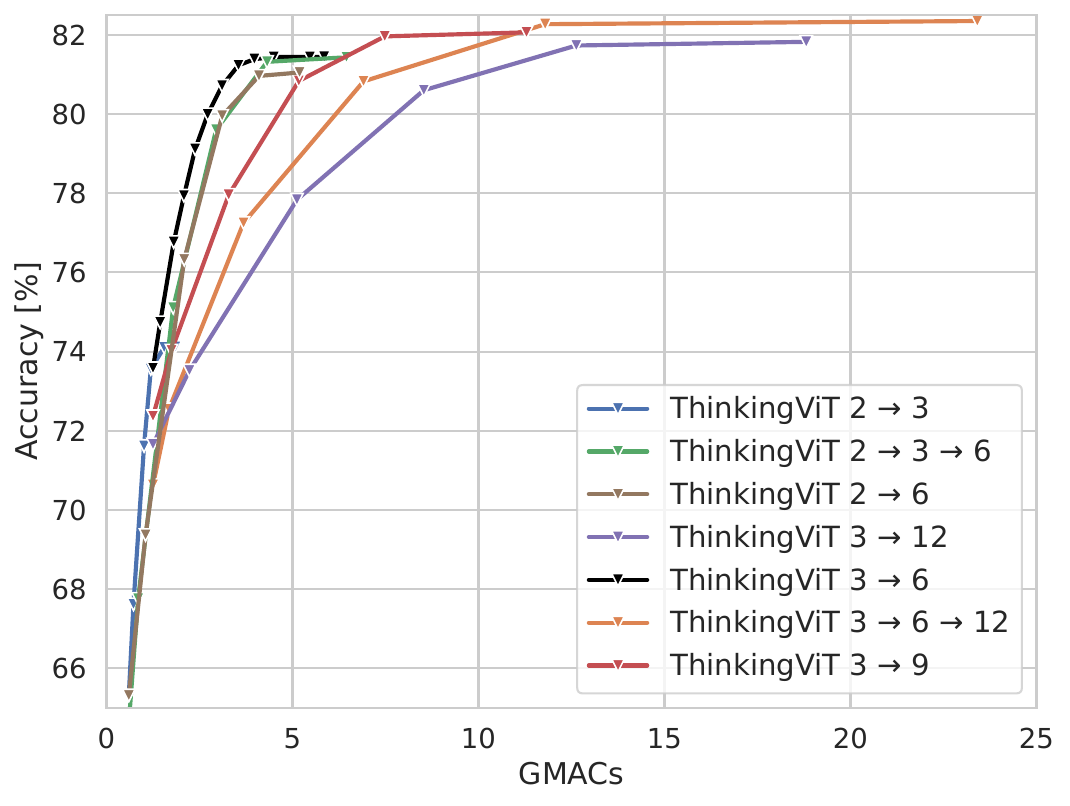}
    \end{subfigure}
    \caption{A higher-resolution version of Figure~\ref{fig:variants}.}
    \label{fig:variantszoomed}
\end{figure*}

Table~\ref{tab:thinking_ablation} analyzes the effect of progressively increasing the attention-head capacity (e.g., 3H$\rightarrow$6H, 3H$\rightarrow$9H, 3H$\rightarrow$12H) and varying loss-weighting schemes between the first and second rounds of thinking.  
Starting with 3 heads and expanding to 6 (3H~$\rightarrow$~6H) yields the best first-round accuracy, while 3H~$\rightarrow$~9H achieves the highest second-round accuracy on ImageNet-1K.
Notably, by using loss weighting we can tune the model to put more focus on the first round or second round based on the desired trade-off between computation and accuracy.
We also explore 3-stage variants in Table~\ref{tab:thinking_depth_ablation}, demonstrating that \name naturally scales to deeper thinking hierarchies and yields higher final accuracy.
See Figure~\ref{fig:variantszoomed} for a high-resolution plot of \name variants.

\begin{table*}[t]
    \centering
    \small
    \caption{Impact of attention expansion and loss weighting on \name with two rounds of progressive thinking on ImageNet-1K.}
    \setlength{\tabcolsep}{1.5pt}
    \begin{tabular}{l c c c}
        \toprule
        \multirow{2}{*}{\textbf{Model Variant}} & 
        \multirow{2}{*}{\textbf{Loss Weight}} & 
        \multicolumn{2}{c}{\textbf{Acc. [\%]}} \\
        &  & \textbf{1\textsuperscript{st} Round} & \textbf{2\textsuperscript{nd} Round} \\
        \toprule
        \name 2H $\rightarrow$ 3H & [0.5, 0.5] & 65.35 & 74.13 \\
        \midrule
        \name 3H $\rightarrow$ 6H & [0.5, 0.5] & 73.58 & 81.44 \\
        \name 3H $\rightarrow$ 6H & [0.4, 0.6] & 73.22 & 81.43 \\
        \name 3H $\rightarrow$ 6H & [0.6, 0.4] & 73.93 & 81.28 \\
        \midrule
        \name 3H $\rightarrow$ 9H & [0.5, 0.5] & 72.51 & 82.02 \\
        \name 3H $\rightarrow$ 9H & [0.4, 0.6] & 71.71 & 82.15 \\
        \name 3H $\rightarrow$ 9H & [0.6, 0.4] & 73.02 & 81.92 \\
        \midrule
        \name 3H $\rightarrow$ 12H & [0.5, 0.5] & 72.03 & 81.70 \\
        \name 3H $\rightarrow$ 12H & [0.4, 0.6] & 70.78 & 81.80 \\
        \name 3H $\rightarrow$ 12H & [0.6, 0.4] & 72.23 & 81.51 \\
        \bottomrule
    \end{tabular}
    \label{tab:thinking_ablation}
\end{table*}

\begin{table*}[t]
\small
\centering
\caption{\name performance with three rounds of thinking on ImageNet-1K, demonstrating that \name naturally scales to deeper thinking hierarchies and yields higher final accuracy.}
\begin{tabular}{l c c c}
    \toprule
\multirow{2}{*}{\textbf{Model}} & \multicolumn{3}{c}{\textbf{Acc. [\%]}} \\
 & \textbf{1\textsuperscript{st} Round} & \textbf{2\textsuperscript{nd} Round} & \textbf{3\textsuperscript{rd} Round} \\
    \midrule
    \name 2H~(33\%)~$\rightarrow$~3H~(50\%)~$\rightarrow$~6H~(100\%) &  64.62& 73.56 & 81.43 \\
    \name 3H~(25\%)~$\rightarrow$~6H~(50\%)~$\rightarrow$~12H~(100\%) & 70.77 & 80.00 & 82.35 \\
    \bottomrule
\end{tabular}
\label{tab:thinking_depth_ablation}
\end{table*}

\end{document}